\documentclass[sigconf]{acmart}

\usepackage{algorithmic}
\usepackage{algorithm}
\usepackage{multirow}

\AtBeginDocument{
  }

\begin{document}
\copyrightyear{2023}
\acmYear{2023}
\setcopyright{acmlicensed}\acmConference[MMAsia '23]{ACM Multimedia Asia 2023}{December 6--8, 2023}{Tainan, Taiwan}
\acmBooktitle{ACM Multimedia Asia 2023 (MMAsia '23), December 6--8, 2023, Tainan, Taiwan}
\acmPrice{15.00}
\acmDOI{10.1145/3595916.3626393}
\acmISBN{979-8-4007-0205-1/23/12}
\title{Block based Adaptive Compressive Sensing with Sampling Rate Control}
 
\author{Kosuke Iwama}
\affiliation{
  \institution{Hosei University}
  \streetaddress{koganei-shi}
  \city{Tokyo}
  \country{Japan}}
  \email{kosuke.iwama.9s@stu.hosei.ac.jp}

\author{Ryugo Morita}
\affiliation{
  \institution{Hosei University}
  \streetaddress{koganei-shi}
  \city{Tokyo}
  \country{Japan}}
  \email{ryugo.morita.7f@stu.hosei.ac.jp}

\author{Jinjia Zhou}
\affiliation{
  \institution{Hosei University}
  \streetaddress{koganei-shi}
  \city{Tokyo}
  \country{Japan}}
   \email{zhou@hosei.ac.jp}

\begin{abstract}
Compressive sensing (CS), acquiring and reconstructing signals below the Nyquist rate, has great potential in image and video acquisition to exploit data redundancy and greatly reduce the amount of sampled data.
To further reduce the sampled data while keeping the video quality, 
this paper explores the temporal redundancy in video CS and proposes a block based adaptive compressive sensing framework with a sampling rate (SR) control strategy.
To avoid redundant compression of non-moving regions, we first incorporate moving block detection between consecutive frames, and only transmit the measurements of moving blocks. The non-moving regions are reconstructed from the previous frame.
In addition, we propose a block storage system and a dynamic threshold to achieve adaptive SR allocation to each frame based on the area of moving regions and target SR for controlling the average SR within the target SR.
Finally, to reduce blocking artifacts and improve reconstruction quality, we adopt a cooperative reconstruction of the moving and non-moving blocks by referring to the measurements of the non-moving blocks from the previous frame. 
Extensive experiments have demonstrated that this work is able to control SR and obtain better performance than existing works.
\end{abstract}



\maketitle
\begin{figure*}
    \includegraphics[width=\textwidth]{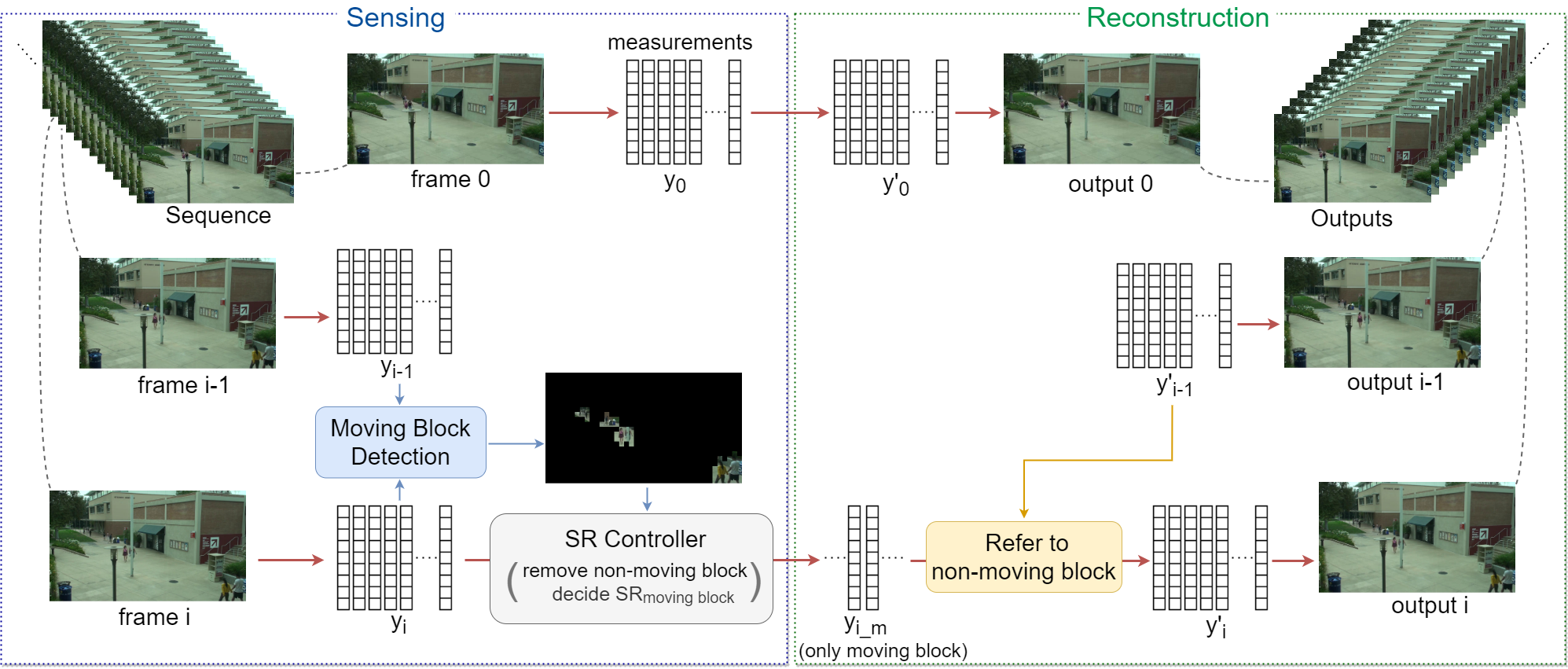}
    \caption{The framework of our proposed method consists of sensing and reconstruction. The sensing part includes measurements-based moving block detection and a sampling rate controller. The reconstruction part includes a method to refer to the non-moving blocks and deep learning based image reconstruction. Each frame is divided into non-overlapping $B \times B$ blocks, and sampling processing is performed block by block.}
    \label{fig:framework}
\end{figure*}

\section{Introduction}
Compressive Sensing (CS) \cite{donoho2006compressed} is a novel sampling and reconstruction method that can recover the original signal from fewer measurements than the Nyquist sampling theorem if the original signal is sparse. 
Mathematically, given an original signal $ x \in \mathbb{R}^N $ , the measurements $ y \in \mathbb{R}^M $ is obtained as follows:
\begin{equation}
  y = \Phi x 
\end{equation}
where $ \Phi \in \mathbb{R}^{M \times N} $ is the sampling matrix, and the sampling rate (SR) is defined as $ M/N (M \ll N) $. 
Due to its potential to greatly enhance the energy efficiency of sensors, it is expected to be applied in various fields such as single-pixel imaging \cite{duarte2008single}, magnetic resonance imaging (MRI) \cite{lustig2007sparse,lustig2008compressed}, and image/video source coding \cite{mun2012dpcm}.

The primary challenge of CS is recovering $x$ from the obtained $y$, which is an uncertain inverse problem to solve.
Many traditional model-driven methods focus on using structural prior distributions with theoretical guarantees, such as sparse representation \cite{elad2010sparse}, low-rank \cite{dong2014compressive}. However, they are computationally expensive and slow, which has presented challenges for practical applications.

Recently, CS performance has been dramatically improved by incorporating a deep convolutional neural network (CNN).
Learning a nonlinear mapping from measurements to the reconstruction image, \cite{kulkarni2016reconnet,shi2019image} improve the recovery accuracy and speed. Also, by incorporating neural networks into the optimization algorithm problem, \cite{zhang2018ista,zhang2020optimization,chen2022fsoinet} further improved the reconstruction quality. Considering human visual attention, \cite{yu2010saliency,zhou2020multi,chen2022content} applied a saliency detection algorithm to achieve adaptive SR assignment to each block.
Although these methods have achieved excellent results for image CS, video CS is still a challenging problem since it requires consideration of both intra-frame and inter-frame correlations.
In addition, video CS is mainly used to capture a scene from a fixed position in real applications. These scenes have many non-moving regions, such as the background. However, many existing methods don't consider this characteristic, leading to redundant compression.

Inspired by this background, the author of \cite{du2021multi} proposed a method to identify the region of interest (ROI) from the difference between the first and subsequent frames and compress only the ROI with high SR. 
But, this method cannot flexibly adapt to changes in the background since the first frame is considered the background. Then, VCSL \cite{yang2023vcsl} introduced the reference frame renewal method into a similar framework and made it possible to adapt to changes in the background.
However, these methods still have three disadvantages: (1) the average SR is not controllable since the SR of each frame depends on the area of the ROI; (2) Correlations between consecutive frames are sometimes disregarded.; (3) The detection accuracy of the moving regions is low.

To solve these problems, we propose a novel video CS method incorporating moving block detection and a SR controller. The proposed method utilizes moving block detection between consecutive frames to cope with background changes and compresses videos more efficiently by reducing the measurements of non-moving blocks. It also introduce a block storage system and dynamic threshold to enable control of the average SR and adaptive SR allocation to each frame based on the area of moving regions and target SR. Furthermore, we adopt a cooperative reconstruction to reduce blocking artifacts and improve reconstruction quality instead of reconstructing the moving and non-moving blocks separately and combining them. The main contributions of this paper are as follows:
 
\begin{itemize}
\item We propose a novel and effective video CS method. By incorporating moving block detection between consecutive frames, it avoids redundant compression of non-moving regions and achieves efficient compression.
\item In order to control the average SR, we propose a block storage system which provides adaptive SR allocation to each frame based on the area of moving regions and target SR. Furthermore, to balance the reconstruction quality and the control of the average SR, we introduce a dynamic threshold method to adjust the threshold appropriately for each frame.
\item
To reduce blocking artifacts and improve reconstruction quality, we adopt a cooperative reconstruction of the non-moving and moving regions.
\item 
Experiments on the VIRAT dataset \cite{oh2011large} demonstrate that our method outperforms state-of-the-art methods in terms of control of the average SR and reconstruction quality.
\end{itemize}

\section{Proposed Method}

\subsection{Overall framework}
According to the existing fixed scene video CS methods \cite{du2021multi,yang2023vcsl}, they required two measurements, one for ROI detection and one for high SR sensing of the ROI. However, in general, actual image sensors allow only one measurement. Therefore, our method measures each frame only once.

As shown in Fig. \ref{fig:framework}, the framework of our proposed method consists of a sensing part including moving block detection and SR controller, and a reconstruction part including a method to refer to the non-moving blocks.
Each frame first obtains measurements $ y_i $ at high SR ($ SR_\text{h} $). 
In the first frame, the measurements $y_0$ is directly transmitted and reconstruction is performed using it.
From the subsequent frame, we first identify moving and non-moving blocks from the previous frame's measurements $ y_{i-1} $ and $ y_i $ by moving block detection module.
Then, to avoid redundant compression of non-moving regions and to control the average SR, SR controller removes the measurements of the non-moving blocks and adjusts the SR of the moving block ($SR_\text{m}$) to obtain the measurements $ y_{i \_ \text{m}} $.
In the reconstruction, $ y_{i \_ \text{m}} $ does not have the measurements of non-moving blocks, so $ y_i' $ is obtained by referring to the measurements of the corresponding block from the previous frame measurements $y_{i-1}'$.
Finally, we perform cooperative reconstruction using $ y_i' $, integrating block gradient descent and proximal mapping, to reduce blocking artifacts and improve the reconstruction quality.

\begin{figure}[h]
  \centering
  \includegraphics[width=\linewidth]{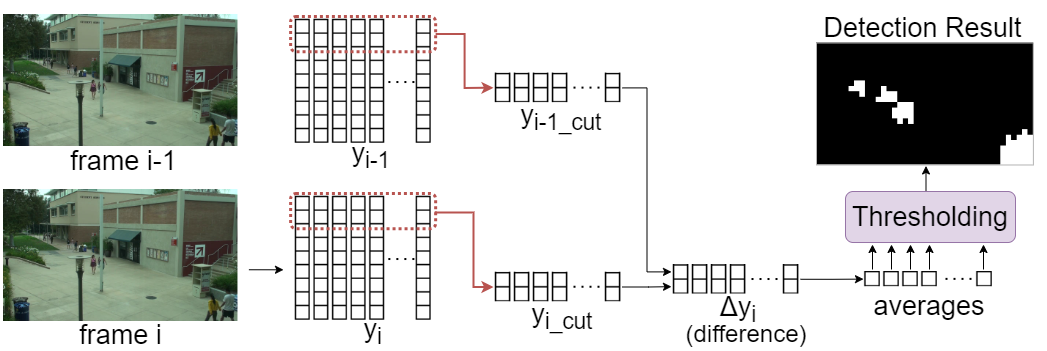}
  \caption{Illustration of Moving Block Detection}
  \label{fig:moving}
\end{figure}

\subsection{Moving Block Detection}
In \cite{du2021multi,yang2023vcsl}, the ROI detection process relies on target and reference frames, but this approach occasionally overlooks inter-frame correlations. To address this issue, we introduce moving block detection between the compressed consecutive frames.

As shown in Fig .\ref{fig:moving}, we use $ y_{i-1} $ and $ y_i $ to identify the moving block in $i$-th frame. First, to compare only low-frequency components, the top portion of each of the measurements is cut off to obtain $ y_{i-1\_\text{cut}} $ and $ y_{i\_\text{cut}} $. Next, the difference between $ y_{i-1\_\text{cut}} $ and $ y_{i\_\text{cut}} $ ($\Delta y_i $) is obtained.
\begin{equation}
  \Delta y_i = | y_{i\_\text{cut}} - y_{i-1\_\text{cut}} |
\end{equation}
Then, the average of each block of $\Delta y_i $ is calculated. If this average is higher than the threshold, it is determined to be a moving block; if it is lower than the threshold, it is determined to be a non-moving block. Based on this information, the measurements of $i$-th frame are reduced and transmitted.

\subsection{SR Control with Dynamic Threshold}
In  \cite{du2021multi,yang2023vcsl}, the average SR depends on the area of the ROI and lacks the ability to control to a certain SR, which poses challenges for real-world applications. To address this limitation, we introduce the SR controller, which considers both the moving regions and the target SR to facilitate its practical application in real-world applications.

Let $ SR_\text{t} $ be the target SR and $n$ the number of frames, then the total SR for each frame is $ n \times SR_\text{t} $. Since the first frame is allocated $ SR_\text{h} $, the total SR available for subsequent frames is $ n \times SR_\text{t} - SR_\text{h} $. Therefore, the control of the average SR is achieved by satisfying the following equation.
\begin{equation}
  \sum_{i=1}^{n-1} SR_i \le n \times SR_\text{t} - SR_\text{h}
  \label{eq:about_SRt}
\end{equation}
Where $SR_i$ is the SR of the $i$-th frame.
To achieve both control of the average SR and reconstruction quality, it is necessary to set SR higher for frames with more moving blocks and lower SR for frames with fewer moving blocks, while satisfying Eq. (\ref{eq:about_SRt}). To accomplish this, we introduce a block storage system and a dynamic threshold. 

\subsubsection{\textbf{Block storage system}}
Setting $ SR_\text{t} $ and $ SR_\text{h} $, the number of moving blocks that are available is obtained from Eq. (\ref{eq:about_SRt}) as follows:
\begin{eqnarray}
  \dfrac{\sum_{i=1}^{n-1} m_i \times SR_\text{h} }{l} = n \times SR_\text{t} - SR_\text{h} \nonumber \\
  \sum_{i=1}^{n-1} m_i = \dfrac{l(n \times SR_\text{t} - SR_\text{h})}{SR_\text{h}}
\end{eqnarray}
where $m_i$ is the number of moving blocks in $i$-th frame and $l$ is the number of total blocks in a frame.

In the block storage system, when there are few moving blocks, the excess is stored in the block storage and allocated to the frame with many moving blocks. This system is set to an initial value, and an additional value to that is added before each frame is measured. The additional value plays a crucial role in preventing block storage from being depleted during the measurement process and maintaining a balanced SR allocation.
On the other hand, the initial value has the role of enabling the system to deal with sequences in which there are many moving blocks in the early frames. The initial value and the additional value satisfy the following equation so that the average SR is equal to the target SR.
\begin{equation}
  \dfrac{SR_\text{h}(b_{\text{ini}}+b_{\text{add}}(n-1))}{l} = n \times SR_\text{t} - SR_\text{h}
\end{equation}
Where $b_{\text{ini}}$ is the initial value and $b_{\text{add}}$ is the additional value.
Given an arbitrary initial value, the additional value is obtained by the following equation.
\begin{equation}
  b_{\text{add}} = \dfrac{l(n \times SR_\text{t} - SR_\text{h}) - b_{\text{ini}} \times SR_\text{h} }{SR_\text{h}(n-1)}
\end{equation}
If the block storage is more than the number of moving blocks, the measurements of the moving block is transmitted as it is, i.e., $ SR_\text{m} = SR_\text{h} $. On the other hand, if the block storage is less than the number of moving blocks, the block storage capacity is insufficient to transmit the measurements as $ SR_\text{m} = SR_\text{h} $. Therefore, to reduce the measurements, $ SR_\text{m} $ is calculated as follows:
\begin{equation}
  SR_\text{m} = \dfrac{SR_\text{h} \times b}{m_i}
\end{equation}
where $b$ is the remaining capacity of the block storage.

\begin{algorithm}[h]
    \caption{Block storage system}
    \label{alg1}
    \begin{algorithmic}[1]    
    \renewcommand{\algorithmicrequire}{\textbf{Input:}}
    \renewcommand{\algorithmicensure}{\textbf{Output:}}
    \renewcommand{\algorithmiccomment}{\ \ \textcolor{blue}{// }}
    \REQUIRE $b_{\text{ini}} , \ b_{\text{add}}, \ m_i $
    \ENSURE $SR_{\text{m}\_i}$ ($SR_\text{m}$ of the $i$-th frame)
    \STATE $b \coloneqq b_{\text{ini}}$
    \STATE $i \coloneqq 1$
    
    \WHILE{$i \le n-1$}
    \STATE $b \coloneqq b+ b_{\text{add}}$ 
    \IF{$m_i \le b$}
    \STATE $SR_\text{m} = SR_\text{h}$
    \STATE $b \coloneqq b- m_i$
    \ELSE
    \STATE $SR_\text{m} = \dfrac{SR_\text{h} \times b}{m_i}$
    \STATE $b = 0$
    \ENDIF
    \STATE $SR_{\text{m}\_i}=SR_\text{m}$  \COMMENT {\textcolor{blue}{output}}
    \STATE $i \coloneqq i + 1$
    \ENDWHILE
    \end{algorithmic}
\end{algorithm}

\begin{table*}[htb]
  \caption{Average PSNR(db)/SSIM comparisons of different video CS methods on VIRAT dataset \cite{oh2011large}. We set $SR_\text{h}$=10.00\% for Multi-rate-VCS and VCSL, and $SR_\text{t}$=1.00\% and $SR_\text{h}$=20.00\% for our method. Ours(w/o DT) denotes our method without the dynamic threshold. The best and second best results are highlighted in red and blue, respectively.}
  \label{tab:sota}
   \begin{tabular}{ccccccccc}
    \toprule
    
    \multirow{2}{*}{sequence} & \multicolumn{2}{c}{Multi-rate-VCS \cite{du2021multi} } &  \multicolumn{2}{c}{VCSL \cite{yang2023vcsl}} & \multicolumn{2}{c}{Ours (w/o DT)} & \multicolumn{2}{c}{Ours} \\
    \cmidrule(lr){2-3} \cmidrule(lr){4-5} \cmidrule(lr){6-7} \cmidrule(lr){8-9} &average SR & PSNR/SSIM &average SR & PSNR/SSIM &average SR & PSNR/SSIM &average SR & PSNR/SSIM \\
    
    \midrule
    S\_0401 & 1.13\% & 33.35/0.9393 & 0.95\% & 34.48/0.9488 & 1.00\% & \textcolor{blue}{34.51}/\textcolor{blue}{0.9494} & 0.99\% & \textcolor{red}{34.62}/\textcolor{red}{0.9505} \\
    S\_0502 & 1.05\% & 31.86/0.9045 & 0.73\% & 34.07/\textcolor{blue}{0.9420} & 0.44\% & \textcolor{blue}{34.29}/0.9349 & 0.96\% & \textcolor{red}{35.41}/\textcolor{red}{0.9494} \\
    S\_0002 & 1.01\% & 35.46/0.9463 & 0.84\% & \textcolor{blue}{36.56}/\textcolor{blue}{0.9539} & 1.00\% & 34.51/0.9276 & 0.98\% & \textcolor{red}{37.06}/\textcolor{red}{0.9548} \\
    S\_0102 & 1.02\% & 34.52/0.9242 & 1.19\% & \textcolor{red}{36.23}/\textcolor{red}{0.9537}	& 1.00\% & 35.81/0.9415 & 0.99\%& \textcolor{blue}{36.17}/\textcolor{blue}{0.9465} \\
    S\_0100 & 0.91\% & 30.61/0.8856 & 0.83\% & \textcolor{blue}{34.25}/\textcolor{blue}{0.9453} & 0.59\% & 34.06/0.9347 & 0.94\% & \textcolor{red}{34.99}/\textcolor{red}{0.9461} \\
    S\_0101 & 1.22\% & 32.47/0.9404 & 0.85\% & \textcolor{blue}{33.55}/\textcolor{red}{0.9533} & 0.52\% & 33.35/0.9438 & 0.98\% & \textcolor{red}{34.18}/\textcolor{red}{0.9533} \\
    average & 1.06\% & 33.05/0.9234 & 0.90\% & \textcolor{blue}{34.86}/\textcolor{blue}{0.9495} & 0.76\% & 34.42/0.9387 & 0.97\% & \textcolor{red}{35.41}/\textcolor{red}{0.9501} \\
    \bottomrule
  \end{tabular}
\end{table*}

\subsubsection{\textbf{Dynamic threshold}}
Although the block storage system enables the control of the average SR, it may not be compatible with high reconstruction quality depending on the threshold. For example, if the threshold is too low, the number of moving blocks will increase, and thus $SR_\text{m}$ will decrease. At the same time, if the threshold is too high, the number of moving blocks will decrease, and thus the accuracy of moving block detection will decrease. Both cases harm reconstruction quality, but it is difficult to avoid this problem if the threshold is fixed since the appropriate threshold varies from frame to frame. Therefore, we solve this problem by using dynamic threshold. Specifically, Our method increases the threshold if $i$-th frame has more moving blocks or less block storage, and decreases the threshold if $i$-th frame has fewer moving blocks or more block storage.

\section{Experiment}
\subsection{Dataset and Implementation Details}
For testing, we select 6 sequences from the public VIRAT-Ground surveillance dataset \cite{oh2011large}. 
All the selected sequences are scene captured from a fixed position, with a resolution of 1280 $\times$ 720.
In the experiments, we set the block size $B$ = 32 and $b_{\text{ini}}$ = 0.5 $\times$ $l$. Regarding the sampling matrix and the reconstruction network, the pre-trained model of CASNet \cite{chen2022content} is selected as the baseline. All experiments are implemented with Pytorch on NVIDIA GeForce RTX 3090.
RTX3090 is only used for applying the deep learning based image reconstruction. The complex of the proposed algorithms including moving block detection and SR controller  is very low, and can be implemented by hardware.

\subsection{Comparison with State-of-the-Art Methods}
\begin{figure}[h]
  \centering
  \includegraphics[width=\linewidth]{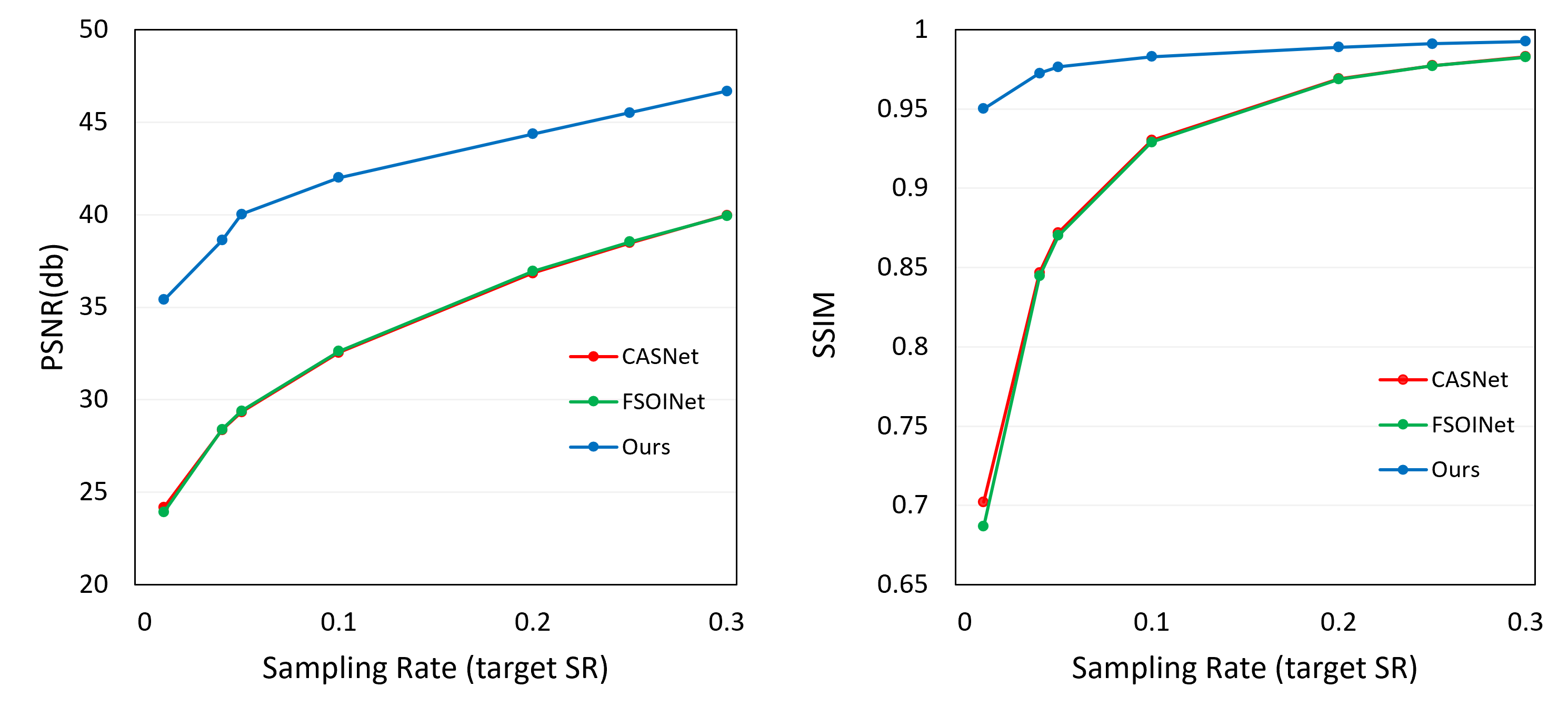}
  \caption{ Comparison of reconstruction performance among state-of-the-art image CS methods CASNet \cite{chen2022content} and FSOINet \cite{chen2022fsoinet} and our method on various target SR. }
  \label{fig:psnr}
\end{figure}

\begin{figure*}[htb]
    \includegraphics[width=\textwidth]{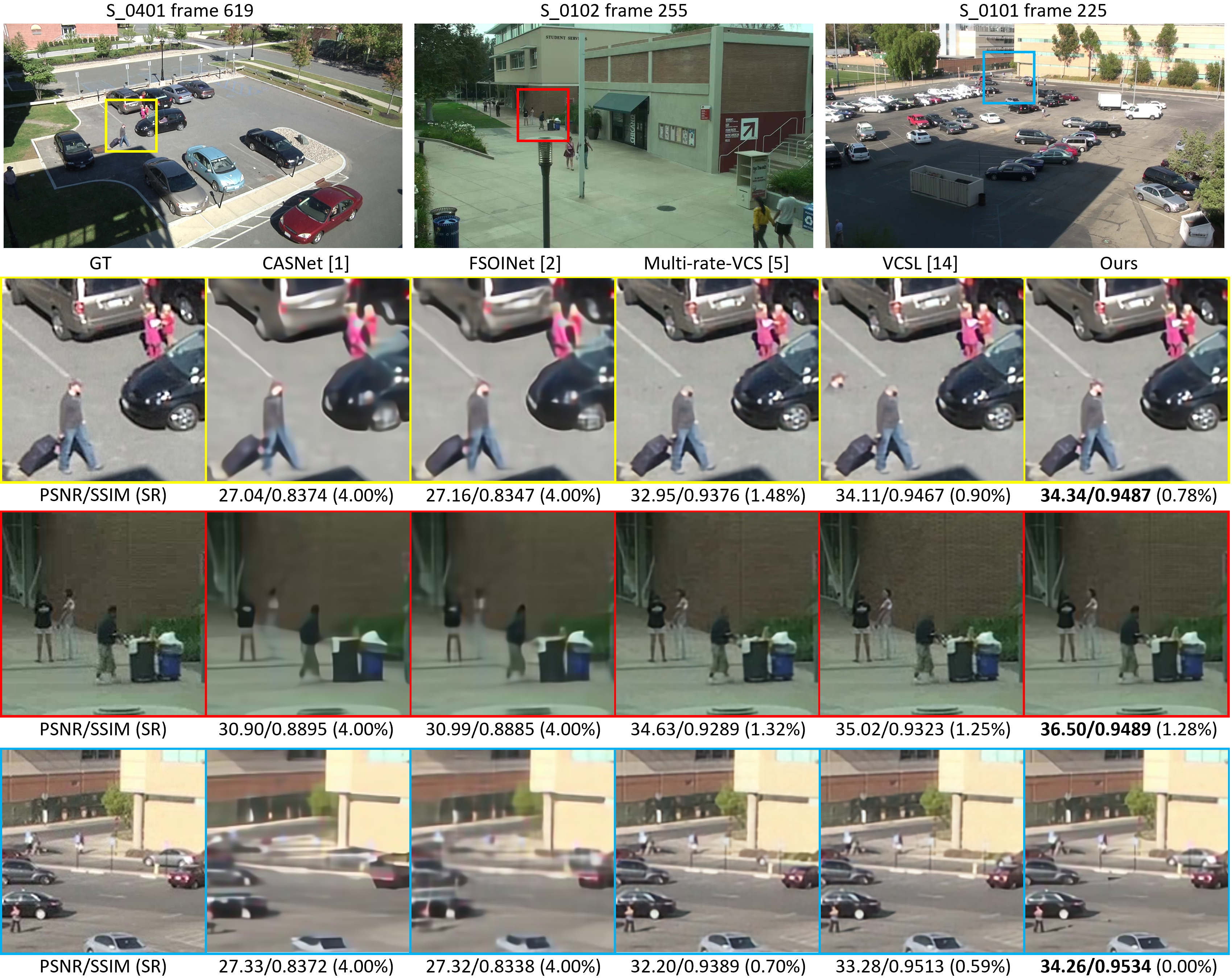}
    \caption{Visual comparison of different CS methods on frame 619 of the 'S\_0401' sequence and frame 255 of the 'S\_0102' sequence and frame 225 of the 'S\_0101' sequence. CASNet and FSOINet are set to SR = 4.00\%, and the other methods are set to the same values as in Table \ref{tab:sota}.}
    \label{fig:visual}
\end{figure*}

We first test all 6 sequences of the dataset and compare our method with state-of-the-art fixed scene video CS methods Multi-rate-VCS \cite{du2021multi} and VCSL \cite{yang2023vcsl}. 
We implement Multi-rate-VCS and VCSL with $SR_\text{h}$ = 10.00\% and observe that their average SR are around 1.00\%, so we set $SR_\text{t}$ = 1.00\% for our method.
As shown in Table \ref{tab:sota}, our method controls the average SR within $SR_\text{t}$ for all sequences, while PSNR and SSIM also surpass the other methods for most sequences. Although there is one sequence in which our method underperforms VCSL in PSNR and SSIM, the average SR of its method is 1.19\%, higher than $SR_\text{t}$, so it is not a fair comparison.
In some sequences, our method exhibits a higher average SR than the other methods, i.e., it is inferior in compression ratio. However, it indicates that our method effectively utilizes SR based on the given $SR_\text{t}$ and achieves high-quality reconstruction, whereas other methods do not fully utilize SR for a given $SR_\text{t}$ and consequently fail to achieve comparable reconstruction quality.
Then, to more extensively evaluate the ability of our method to control the average SR and the reconstruction quality, we test all 6 sequences at various $SR_\text{t}$ and compare it to the state-of-the-art image CS methods CASNet \cite{chen2022content} and FSOINet \cite{chen2022fsoinet}. Fig. \ref{fig:psnr} shows the transition curves of average PSNR and SSIM for each method. From this result, it is evident that our method outperforms existing methods in all $SR_\text{t}$. Additionally, Table \ref{tab:average SR} presents the results of the average SR for each sequence in our method at various $SR_\text{t}$, indicating that the average SR is effectively controlled based on the given $SR_\text{t}$. These findings demonstrate that our method is adaptable to various $SR_\text{t}$ and provides superior reconstruction quality.
Fig. \ref{fig:visual} presents a visual comparison of reconstruction images between our method and existing methods. CASNet and FSOINet exhibit a significant decline in image quality, while Multi-rate-VCS and VCSL show missing moving regions. In contrast, our proposed method exhibits high-quality reconstructions with no loss of moving regions. It demonstrates the efficacy of our method in accurately detecting moving regions and surpassing the other methods from a visual perspective.

\begin{table}[h]
  \caption{The average SR of our method for each sequence at various target SR.}
  \label{tab:average SR}
   \begin{tabular}{ccccccc}
    \toprule
    
    \multirow{2}{*}{sequence} & \multicolumn{6}{c}{target SR} \\
    
    \cmidrule{2-7} & 4.00\% & 5.00\% & 10.00\% & 20.00\% & 25.00\% & 30.00\%\\
    \midrule
    S\_0401 & 3.91\% & 4.90\% & 9.86\% & 19.84\% & 24.63\% & 29.51\% \\
    S\_0502 & 3.91\% & 4.84\% & 9.72\% & 19.43\% & 24.24\% & 29.03\% \\
    S\_0002 & 3.96\% & 4.94\% & 9.89\% & 19.88\% & 24.87\% & 29.55\% \\
    S\_0102 & 3.85\% & 4.73\% & 8.36\% & 17.45\% & 22.07\% & 26.60\% \\
    S\_0100 & 3.69\% & 4.50\% & 8.44\% & 17.42\% & 22.05\% & 26.48\% \\
    S\_0101 & 3.89\% & 4.81\% & 9.58\% & 19.03\% & 23.71\% & 28.34\% \\
    average & 3.87\% & 4.78\% & 9.31\% & 18.84\% & 23.59\% & 28.25\% \\
    \bottomrule
  \end{tabular}
\end{table}

\subsection{Analysis and Discussion}
In this section, we mainly validate the effectiveness of the block storage system and dynamic threshold. For more in-depth evaluation, we first conduct two experiments using three different approaches: without the block storage system, without dynamic threshold, and with both (ours). In one experiment, the initial threshold is set to 0.04 and the three approaches are tested at multiple $SR_\text{t}$. In the other experiment, $SR_\text{t}$ is fixed at 1.00\% and the three approaches are tested at multiple initial thresholds. The quantitative results for each experiment are shown in Table \ref{tab:ablation} and Table \ref{tab:ablation2}, respectively.

When the block storage system is not utilized, it is observed that the average SR cannot be controlled based on $SR_\text{t}$ and the average SR is dependent on the threshold; lower threshold lead to higher average SR, while higher threshold result in lower average SR. In contrast, the approach using the block storage system allows for controlling the average SR within $SR_\text{t}$ for any initial threshold and $SR_\text{t}$ combination.

\begin{table*}[htb]
  \caption{Analysis of block storage system (BSS) and dynamic threshold (DT) for multiple target SR on 'S\_0401' sequence. We set the initial threshold to 0.04.}
  \label{tab:ablation}
   \begin{tabular}{cccccccccc}
    \toprule
    \multicolumn{2}{c}{target SR} & \multicolumn{2}{c}{0.80\% } &  \multicolumn{2}{c}{0.90\%} & \multicolumn{2}{c}{1.00\%} & \multicolumn{2}{c}{1.10\%} \\
     \cmidrule(lr){3-4} \cmidrule(lr){5-6} \cmidrule(lr){7-8} \cmidrule(lr){9-10} BSS & DT & average SR & PSNR/SSIM &average SR & PSNR/SSIM &average SR & PSNR/SSIM & average SR & PSNR/SSIM\\
    
    \midrule
    $\times$ & $\times$ & 1.03\% 
& 34.69/0.9511 & 1.03\% & 34.69/0.9511 & 1.03\% & 34.69/0.9511 & 1.03\% & 34.69/0.9511 \\
    \checkmark & $\times$ & 0.80\% 
& 32.42/0.9242 & 0.90\% & 33.62/0.9392 & 1.00\% & 34.51/0.9494 & 1.03\% & 34.69/0.9511 \\
\checkmark & \checkmark & 0.79\% 
& 34.26/0.9482 & 0.89\% & 34.43/0.9494 & 0.99\% & 34.57/0.9503 & 1.09\% & 34.72/0.9513 \\
    
    \bottomrule
  \end{tabular}
\end{table*}

\begin{table*}[htb]
  \caption{Analysis of block storage system (BSS) and dynamic threshold (DT) for multiple initial thresholds on 'S\_0401' sequence. We set $SR_\text{t}$=1.00\%.}
  \label{tab:ablation2}
   \begin{tabular}{cccccccccc}
    \toprule
    \multicolumn{2}{c}{threshold} & \multicolumn{2}{c}{0.020 } &  \multicolumn{2}{c}{0.030} & \multicolumn{2}{c}{0.040} & \multicolumn{2}{c}{0.050} \\
     \cmidrule(lr){3-4} \cmidrule(lr){5-6} \cmidrule(lr){7-8} \cmidrule(lr){9-10} BSS & DT & average SR & PSNR/SSIM &average SR & PSNR/SSIM &average SR & PSNR/SSIM & average SR & PSNR/SSIM\\
    
    \midrule
    $\times$ & $\times$ & 6.18\% & 35.53/0.9627 & 3.32\% & 35.29/0.9584 & 1.03\% & 34.69/0.9511 & 0.57\% & 33.71/9448\\
    \checkmark & $\times$ & 1.00\% & 24.38/0.7435 & 1.00\% & 26.65/0.8247 & 1.00\% & 34.51/0.9494 & 0.57\% & 33.71/9448\\
\checkmark & \checkmark & 0.99\% & 34.62/0.9505 & 0.99\% & 34.62/0.9505 & 0.99\% & 34.62/0.9505 & 0.99\% & 34.62/0.9505\\
    
    \bottomrule
  \end{tabular}
\end{table*}

\begin{figure}[htb]
  \centering
  \includegraphics[width=\linewidth]{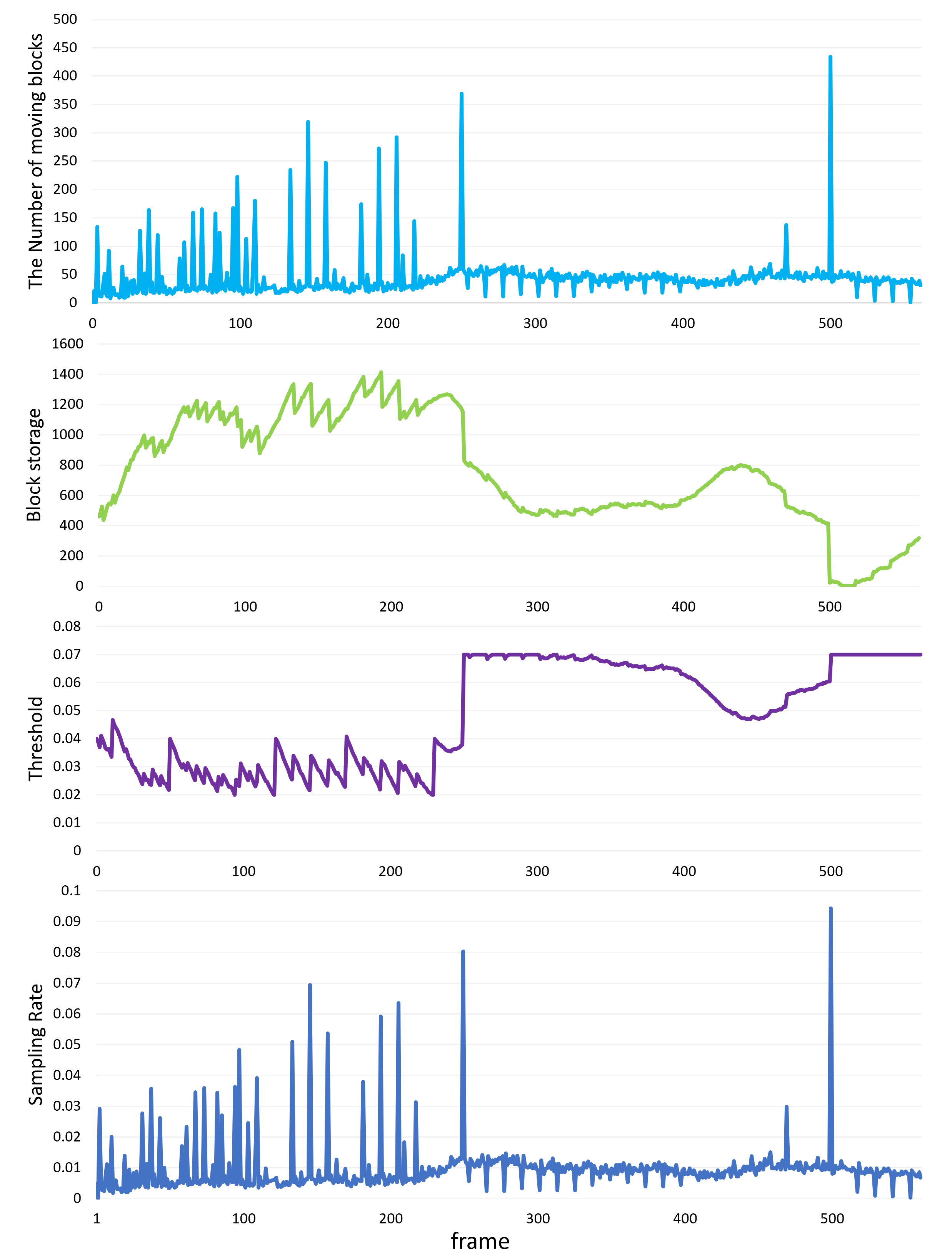}
  \caption{The transition curves for the number of moving blocks, block storage, threshold, and sampling rate, respectively on 'S\_0102' sequence at $SR_{\text{t}}$ = 1.00\%.}
  \label{fig:stb}
\end{figure}
However, it is observed that without the introduction of dynamic threshold, the reconstruction quality degrades. Especially when the thresholds deviate significantly from the appropriate values, the quality suffers considerably. In contrast, our method consistently delivers high-quality reconstructions regardless of the initial threshold. It implies that the dynamic threshold plays a crucial role in preventing the lowering of $SR_\text{m}$ and the degradation of moving block detection accuracy, while effectively optimizing the reconstruction quality within $SR_\text{t}$.

For a more detailed analysis, we visualize the transition of each value in our method for the 'S\_0102' sequence, shown in Fig. \ref{fig:stb}. 
It can be seen that frames with fewer moving blocks are assigned a lower SR, leading to an increase in block storage, and conversely, frames with a higher number of moving blocks are assigned a higher SR, leading to a decrease in block storage.
It indicates that our method achieves adaptive SR allocation to each frame based on the number of moving blocks, facilitated by the block storage system. Furthermore, it can be observed that the threshold vary depending on the number of moving blocks and block storage. This suggests an attempt to optimize the reconstruction quality within the given target SR.

\section{Conclusion}
In this paper, we propose a novel video CS method that collectively achieves efficient compression considering non-moving regions, control of the average SR, and high-quality reconstruction. By incorporating moving block detection between consecutive frames, it accurately identifies the moving regions and removes data in the non-moving regions, which avoids redundant compression in the non-moving regions. Furthermore, we introduce a block storage system and dynamic threshold to allocate adaptive SR to each frame based on the area of the moving region and the target SR, thus satisfying both the control of the average SR and the reconstruction quality. Moreover, we adopt cooperative reconstruction of the non-moving and moving blocks to reduce blocking artifacts and improve reconstruction quality. Experimental results demonstrate that our proposed method is quantitatively and qualitatively superior to the state-of-the-art methods while controlling the average SR. This advancement has promising implications for enhancing the efficiency and effectiveness of video CS.
 
\begin{acks}
This work was supported by JSPS KAKENHI Grant Number JP22K12101.
\end{acks}

\bibliographystyle{ACM-Reference-Format}
\bibliography{base}

\end{document}